\documentclass[letterpaper, 10 pt, conference]{ieeeconf}
\IEEEoverridecommandlockouts 
\usepackage[letterpaper, left=52pt, right=52pt, bottom=44pt, top=57pt]{geometry}

\usepackage{graphicx}
\usepackage{amsmath} 
\usepackage{amssymb}  
\usepackage{mathtools}
\usepackage{bbm}
\usepackage{multirow}
\usepackage[dvipsnames]{xcolor}
\usepackage{subfig}
\usepackage[noadjust]{cite}
\usepackage{enumerate}
\usepackage{booktabs}

\usepackage{algpseudocode}
\usepackage{algorithm}
\usepackage{caption}
\newcommand*\concat{\mathbin{\|}}

\title{\LARGE \bf
Long-term Pedestrian Trajectory Prediction \protect\\ using Mutable Intention Filter and Warp LSTM}

\author{Zhe Huang, Aamir Hasan, Kazuki Shin, Ruohua Li, and Katherine Driggs-Campbell
\thanks{Z. Huang, A. Hasan, K. Shin, R. Li, and K. Driggs-Campbell are with the Department of Electrical and Computer Engineering at the University of Illinois at Urbana-Champaign. email: \{zheh4, aamirh2, kazukis2, ruohual2, krdc\}@illinois.edu}
}

\begin{document}

\maketitle
\thispagestyle{empty}
\pagestyle{empty}

\begin{abstract}
Trajectory prediction is one of the key capabilities for robots to safely navigate and interact with pedestrians. Critical insights from human intention and behavioral patterns need to be integrated to effectively forecast long-term pedestrian behavior. Thus, we propose a framework incorporating a Mutable Intention Filter and a Warp LSTM (MIF-WLSTM) to simultaneously estimate human intention and perform trajectory prediction. The Mutable Intention Filter is inspired by particle filtering and genetic algorithms, where particles represent intention hypotheses that can be mutated throughout the pedestrian's motion. Instead of predicting sequential displacement over time, our Warp LSTM learns to generate offsets on a full trajectory predicted by a nominal intention-aware linear model, which considers the intention hypotheses during filtering process. Through experiments on a publicly available dataset, we show that our method outperforms baseline approaches and demonstrate the robust performance of our method under abnormal intention-changing scenarios. Code is available at \texttt{https://github.com/tedhuang96/mifwlstm}.
\end{abstract}

\section{Introduction}\label{sec:Introduction}
As humans, we make effective predictions of pedestrian trajectories over long time horizons even when novel behavior is present~\cite{baldwin2001discerning}, and continually estimate people's underlying goals or intent from subtle motion~\cite{blake2007perception}. If robots are to similarly navigate and interact safely with people, reliable forecasts of human behavior are required~\cite{rudenko2019human}. For long-term prediction of pedestrian trajectories, two key challenges are posed for robots to emulate human performance. The first challenge is to reason about human intention given the observation history of pedestrian motion. The second challenge is to build a generative model that captures social norms and takes full advantage of intention information.

Pioneering work tackled the intention inference challenge by treating it as either a classification problem \cite{volz2016data, xue2017bi, ghori2018learning, rasouli2019pie} or a filtering problem~\cite{rehder2015goal, particke2018improvements, du2019online}. Model-based \cite{helbing1995social, van2011reciprocal, yamaguchi2011you, kuderer2012feature} and learning-based \cite{alahi2016social, gupta2018social, zhang2019sr, fernando2018soft+} techniques are applied to modeling pedestrian behavior. While these previous studies have made significant progress, we see several potential limitations nonetheless. First, using a short sliding window of trajectory data as the input to prediction model is a typical choice for learning-based methods \cite{alahi2016social}. The prediction results from the last sliding window may not be used to assist the prediction on the current sliding window, which leads to waste of previous observation history. Second, the influence on trajectory prediction from human intention and pedestrian motion pattern is nontrivial to balance. Model-based methods require massive parameter tuning and the performance may vary across pedestrians \cite{ferrer2014behavior}. Learning-based methods, such as recurrent neural network (RNN), predict the displacement between pedestrian positions in neighboring frames. Though displacement is easier to predict than global position, error accumulation on sequential displacement prediction typically results in drift of long-term pedestrian trajectory prediction~\cite{becker2018red}. We find this drifting issue exists even if the desired goal is an input to the RNN.

\begin{figure*}[bht!]
	\centering
	\includegraphics[width=175mm]{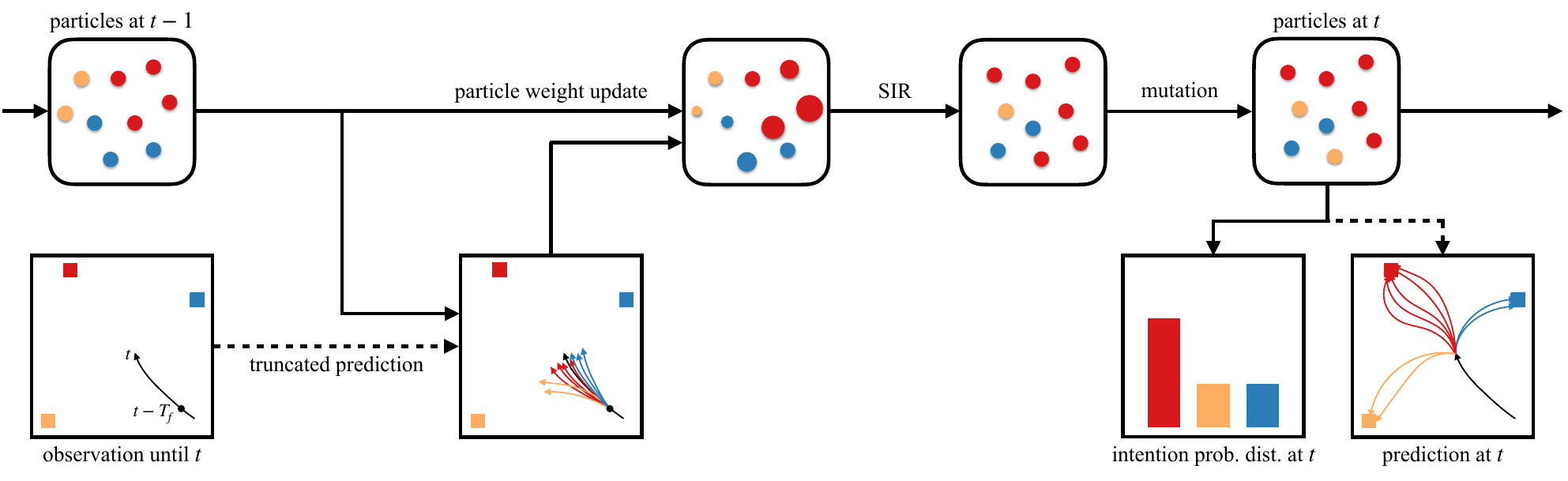}
	\caption{Overview of our trajectory prediction framework MIF-WLSTM. WLSTM (dashed-line arrow) performs both truncated prediction for particle weight update and long-term prediction after mutation. MIF takes truncated prediction results to update particle weights, and implements sequential importance resampling (SIR) and intention mutation mechanism (IMM). Finally, the framework outputs a belief over intentions and long-term prediction samples at $t$.}
    \label{fig:intention_filter}
    \vspace{-5pt}
\end{figure*}

To overcome these limitations, we propose a trajectory prediction framework (Fig. \ref{fig:intention_filter}), which fuses a Mutable Intention Filter (MIF) and a Warp LSTM (WLSTM). MIF performs particle filtering to make full use of the whole observation history, and to maintain a belief over all potential desired goal regions (i.e., intention hypotheses). A novel intention mutation mechanism is introduced from genetic algorithms to make the prediction framework resilient to premature convergence on intentions and intention-changing motions. As the motion model of our filter, WLSTM simultaneously generates offsets over a full trajectory to avoid the accumulative error dilemma, where the trajectory is predicted by a heuristic method that embeds the intention information. WLSTM adds the offset to warp the nominal trajectory with the aim of capturing influences from the map (e.g., avoid obstacles) and from human tendencies (e.g., nonlinear behavior). Our contributions are fourfold:


\begin{enumerate}[1.]
    \item A novel framework is presented to simultaneously estimate pedestrian intentions and generate long-term trajectory prediction samples with a flexible sampling strategy.
    \item A Mutable Intention Filter inspired from genetic algorithms is proposed to robustly perform intention estimation even for abnormal pedestrian behavior.
    \item A Warp LSTM is introduced to predict offsets on an intention-aware nominal prediction to capture pedestrian behavior over long time horizons.
    \item We apply a bidirectional structure across both observation and prediction period to propagate physical intention information through the whole trajectory.
\end{enumerate}

This paper is organized as follows. Section \ref{sec:2} summarizes learning-based methods for pedestrian trajectory prediction and the methodology related to our work. Section \ref{sec:3} formulates the problem, and describes MIF and WLSTM in details. Section \ref{sec:4} elaborates on experiments and a parametric study of our framework MIF-WLSTM. In addition, Section \ref{sec:4} demonstrates that our method surpasses baselines on a publicly available dataset with trajectory visualization and quantitative evaluation. Our conclusions and future work are presented in Section \ref{sec:5}.

\section{Related Work}\label{sec:2}

\textbf{Intention Filter.} In practice, trajectory data is recorded once the pedestrian is detected. Information from previously recorded data is essential to infer key properties of long-term pedestrian motion such as the intention \cite{ziebart2009planning, kitani2012activity}. Kalman filter based on interacting multiple models \cite{schneider2013pedestrian} takes into account different pedestrian motion types and has been applied to pedestrian intention recognition \cite{schulz2015controlled}. Particle filter is an alternative filtering approach that uses particles (i.e., samples) to model various distributions over pedestrian goals besides the Gaussian distribution \cite{rehder2015goal, particke2018improvements, du2019online}. Many of these filtering approaches do not consider premature convergence to wrong intentions, while some work tried to alleviate this issue by weighting more on less probable intentions \cite{hug2018particle}. Our MIF guarantees the diversity among all intention hypotheses throughout the filtering process, so the premature convergence can always be recovered.

\textbf{RNN for Pedestrian Trajectory Prediction.} The power of RNNs in generating sequences is used to model pedestrian behavior from various perspectives \cite{graves2013generating}. For short-term human-human interaction, many socially aware structures~\cite{alahi2016social, vemula2018social, gupta2018social, ivanovic2019trajectron} are proposed to encode hidden states of neighbors through an RNN from observed trajectories. Many research efforts focus on integrating contextual cues into RNNs for long-term trajectory prediction, including intent and map information \cite{rudenko2019human, karasev2016intent}. Convolutional Neural Networks are used to extract map information from scene images~\cite{sadeghian2019sophie} or high-definition semantic maps \cite{salzmann2020trajectron++}. The distance from humans to static obstacles may be encoded to introduce the influence from the obstacles to the humans~\cite{bartoli2018context}. As for intent, the probability distribution over possible goal regions may be used to select among RNNs trained for different intentions~\cite{xue2017bi}, or used as input to the downstream of the architecture~\cite{katyalintent, rasouli2019pie}. Most of these methods treat the information during observation period as input and outputs the trajectory during prediction period, which is represented as an input-to-output architecture \cite{zhang2019sr}. In contrast, we unroll intention information by predicting an intention-aware nominal trajectory. A bidirectional RNN is implemented across the whole trajectory concatenated between observation and nominal prediction, so as to efficiently pass the intention message through the whole time period. Though bidirectional structure was proposed in the past \cite{xue2017bi, saleh2018cyclist}, they applied the structure across only observation data, and did not break the ``wall'' between observation and prediction period.



\textbf{Residual Structure for LSTM.} Shortcut connection between network layers builds a gradient highway to help neural networks (e.g., RNNs, extremely deep CNNs) learn effectively through back propagation~\cite{he2016deep}. 
LSTM itself provides an uninterrupted gradient flow between cell states in temporal domain to alleviate vanishing or exploding gradient problems~\cite{hochreiter1997long}. Stacked residual LSTM is proposed for phrase generation tasks, where shortcut paths are added between LSTM layers and attempts to achieve efficient training on deep LSTM networks~\cite{prakash2016neural}. Similar structures are explored in various sequential tasks \cite{kim2017residual, wang2018using, wang2016recurrent, zhao2018deep, kim2018residual}. Our WLSTM uses the nominal trajectory as input, and outputs offset that is added to the input trajectory in order to ``warp'' towards the ground truth. This warping action resembles the residual learning architecture, which adds the learned residual to the input gradient flow. We regard our method as a variant of Residual LSTM, which is specifically structured for learning physical offset rather than gradient flow.






\section{Method}\label{sec:3}

In this work, we assume that a human has an unknown intention $G$, which denotes a desired goal region. The final position of human trajectory $g$ is located in $G$. $G$ belongs to a finite set of potential intentions $\mathcal{G} = \{G_1, G_2, \ldots, G_m\}$, which is assumed prior map knowledge \cite{bandyopadhyay2013intention}. At timestamp $t$, the observation history of the human's positions $x_{1:t}$ is available. Our goal is to concurrently infer $G$ from observations and predict future positions $x_{t+1:t+T_f}$. $x_t \in \mathbb{R}^2$ denotes a global position and $T_f$ represents a lookahead time window.

Trajectory prediction is divided into two problems corresponding to the two challenges in Section \ref{sec:Introduction}. The first problem is how to correctly estimate the intention online based on the recorded trajectory? The second problem is given the observation and the intention estimate, how to effectively predict human trajectory? We introduce MIF and WLSTM to simultaneously solve both problems. Fig. \ref{fig:intention_filter} shows the architecture of the entire framework integrating both methods.
The MIF requires a motion model to update the belief on potential intentions, while the WLSTM predicts trajectory using the estimated intention from the filtering process. We will show in Section \ref{sec:4} that the more accurate motion modeling from WLSTM introduces less noise to intention inference, and the intention robustly estimated from the MIF offers valuable information to trajectory prediction.

\subsection{Mutable Intention Filter}

As pedestrian trajectory prediction is a multi-modal problem~\cite{gupta2018social}, we propose MIF (see Algorithm \ref{algo:MIF}) that applies particle filtering to generating multiple prediction samples with different hypotheses on pedestrian intentions. Moreover, the MIF yields a belief over potential intentions $b_G$ that converges to the correct intention, even if the pedestrian changes its intention during motion.

\begin{algorithm}[t!]
    \caption{Mutable Intention Filter}\label{algo:MIF}
    \begin{algorithmic}

    \State \textbf{Input:} $x_{1:t}, G_{1:m}, f$
    \State \textbf{Output:} $b_{G}, X_{pred}$ during each iteration
    \For {$\omega^{(i)}, G^{(i)}$ in $\Omega, \mathcal{G}$ } \Comment{initialize particles}
        \State $\omega^{(i)} \gets 1/M$
        \State $G^{(i)} \gets$ \Call{IntentionSampling}{$G_{1:m}$}
    \EndFor
    \For {each iteration} \Comment{filtering}
        \State $X_{truncated} \gets \varnothing, X_{pred} \gets \varnothing$
        \For {$i = 1$ to $M$}
            \State $\hat{x}^{(i)}_{t-T_{f}+1:t+T_g^{(i)}} \gets$
            \State $\qquad\Call{TrajectoryPrediction}{f, x_{1:t-T_{f}}, G^{(i)}}$
            \State $X_{truncated} \gets \Call{Append}{X_{truncated}, \hat{x}^{(i)}_{t-T_{f}+1:t}}$
        \EndFor
        \State $\Omega \gets \Call{WeightUpdate}{\Omega, X_{truncated}, x_{t-T_{f}+1:t}}$
        \State $\Omega, \mathcal{G} \gets \Call{SequentialImportanceResampling}{\Omega, \mathcal{G}}$
        \State $\mathcal{G} \gets \Call{IntentionMutation}{\mathcal{G}}$
        \State $b_{G} \gets \Call{IntentionAggregation}{\Omega, \mathcal{G}}$
        \For {$i = 1$ to $M$}
            \State $\hat{x}^{(i)}_{t+1:t+T_g^{(i)}} \gets$
            \State $\qquad\Call{TrajectoryPrediction}{f, x_{1:t}, G^{(i)}}$
            \State $X_{pred} \gets \Call{Append}{X_{pred}, \hat{x}^{(i)}_{t+1:t+T_g^{(i)}}}$
        \EndFor
        \State \textbf{yield} $b_{G}, X_{pred}$
        \State $x_{1:t+1} \gets \Call{Append}{x_{1:t}, x_{t+1}}$ \Comment{new observation}
    \EndFor
 \end{algorithmic} 
 \end{algorithm}
 
The initialization of MIF is activated when a pedestrian is detected. $M$ particles are initialized with normalized uniform weights $\Omega = \{\omega^{(1)} \ldots, \omega^{(M)}\}$ and with intention hypotheses $\mathcal{G} = \{G^{(1)}, G^{(2)}, \ldots, G^{(M)}\}$ uniformly sampled from $G_{1:m}$. To generate an intention-aware prediction sample from the $i$th particle, a goal position $g^{(i)}$ is randomly sampled from the goal region hypothesis $G^{(i)}$. The average displacement magnitude is computed based on history $x_{1:t}$, and divides the distance between $x_t$ and $g^{(i)}$ to heuristically obtain the remaining time step $T_g$. $x_{1:t}$, $T_g$, and $g$ are fed into a motion model $f$ to perform trajectory prediction. 




\begin{algorithm}[t!]
    \caption{Intention-aware Trajectory Prediction}\label{algo:ITP}
    \begin{algorithmic}
    \Function{TrajectoryPrediction}{$f, x_{1:t}, \hat{G}$}
        \State $g \gets$ \Call{GoalPositionSampling}{$\hat{G}$}
        \State $T_{g} \gets$ \Call{TimeToGoHeuristic}{$x_{1:t}, g$}
        \State $\hat{x}_{t+1:t+T_g} \gets f\left(x_{1:t}, g, T_{g}\right)$
        \State \Return $\hat{x}_{t+1:t+T_g}$
    \EndFunction
\end{algorithmic} 
\end{algorithm}

\begin{algorithm}[t!]
    \caption{Intention Mutation Mechanism}\label{algo:IMM}
    \begin{algorithmic}
    \Function{IntetionMutation}{$\mathcal{G}$}
        \State $\mathcal{G}_{new} \gets \varnothing$
        \For {$G^{(i)}$ in $\mathcal{G}$}
            \State $ p \sim U(0, 1)$
            \If {$p < p_{mutation}$}
                \State $G^{(i)} \gets \Call{IntentionSampling}{G_{1:m} \setminus \{G^{(i)}\}}$
            \EndIf
            \State $\mathcal{G}_{new} \gets \Call{Append}{\mathcal{G}_{new}, G^{(i)}}$
        \EndFor
        \State \Return $\mathcal{G}_{new}$
    \EndFunction
\end{algorithmic} 
\end{algorithm}

During the filtering iteration at time $t$ and given observation $x_{1:t}$, we treat $x_{1:t-T_{f}}$ as the input and $x_{t-T_{f}+1:t}$ as the ground truth. Trajectory prediction is implemented across all particles, and prediction samples $\hat{x}^{(i)}_{t-T_f+1:t+T_g^{(i)}}$ are truncated at timestamp $t$ to be compared against the ground truth. We update the weight $\omega_i$ based on L2 distance between the $i$th prediction and the ground truth:

\begin{equation}
\omega^{(i)}\leftarrow\frac{\omega^{(i)} \exp{\left(-\tau ||\hat{x}^{(i)}_{t-T_{f}+1:t} - x_{t-T_{f}+1:t}||\right)}}{\sum\limits_{i=1}^M \omega^{(i)} \exp{\left(-\tau ||\hat{x}^{(i)}_{t-T_{f}+1:t} - x_{t-T_{f}+1:t}||\right)}}
\end{equation}

\noindent where $\tau$ is a temperature hyperparameter that tunes exploration and exploitation among intention hypotheses. Lower deviation leads to larger weight during the weight update step.

Sequential importance resampling (SIR) is implemented after the weight update to avoid sample degeneracy \cite{liu2001theoretical}. Particles are resampled based on updated weights, and the intention hypotheses are inherited from the last generation. The weights of particles in new generation are again uniform. The number of particles is fixed throughout the resampling process. In order to prevent premature convergence and to address intention-changing cases, an intention mutation mechanism (IMM) inspired by genetic algorithms follows SIR. Essentially, the inherited intention $G^{(i)}$ has a tiny possibility of mutating to a different intention, which imitates the scenario when a pedestrian changes its destination midway through the trajectory. We demonstrate in Section \ref{sec:4} that the mutation mechanism enables the intention filter to adaptively predict pedestrian trajectories under intention-changing scenarios.


At the end of the iteration at $t$, $\omega^{i}$'s are aggregated in terms of intention hypotheses to yield the belief $b_G$ over all possible intentions. Furthermore, the whole observation $x_{1:t}$, updated $G^{(i)}$, and the motion model $f$ are used to output multi-modal long-term trajectory prediction samples $\hat{x}^{(i)}_{t+1:t+T_g^{(i)}}$.

\subsection{Warp LSTM}

When applying LSTM to trajectory prediction, it is a common practice to predict position displacement between neighboring time steps in place of global positions~\cite{graves2013generating, xu2018encoding}. Trajectory data is transformed into a more standardized format that is easier to learn for LSTM. We develop another standardization concept named offset. The offset is defined as the difference between human's trajectory and nominal prediction at each time step. While predicted displacements have to be fed recurrently to predict counterparts at following time steps, offsets along the whole trajectory can be predicted in parallel. Besides the faster computation speed, offset prediction also addresses the drifting prediction issue caused by error accumulation in sequential prediction, since offset at next time step does not depend on the previous offsets.


To effectively learn the offset due to physical constraints and personal preferences~\cite{sadeghian2019sophie, salzmann2020trajectron++, carvalho2019long}, we first introduce an intention-aware linear model (ILM) to capture the effect of intention on trajectory prediction. We assume that during training stage, the ground truth goal position $g$ and the remaining time step $T_g$ to reach goal are known, whereas during evaluation stage, the intention hypotheses $G^{(i)}$'s from the MIF will provide estimates on both (see Algorithm \ref{algo:ITP}). Given $g$ and $T_g$, ILM creates a straight line $\Tilde{x}_{t+1:t+T_g}$ connecting the current position $x_t$ and $g$, which reflects the fact that people attempt to reach their desired goals with minimum effort \cite{dragan2013legibility}. The observation $x_{1:t}$ and the nominal prediction $\Tilde{x}_{t+1:t+T_g}$ are concatenated as a nominal full trajectory.

\begin{equation}
\hat{x}_{1:t+T_g} = \left[x_{1:t} \concat \Tilde{x}_{t+1:t+T_g}\right]
\end{equation}

WLSTM receives the nominal trajectory $\hat{x}_{1:t+T_g}$ as input, and attempts to learn the offset to warp $\hat{x}_{1:t+T_g}$ towards the ground truth $x_{1:t+T_g}$. We find this offset learning naturally fits the residual learning procedure. Ideally, we desire an underlying mapping $\mathcal{H}(\hat{x})$ to map an unimpressive nominal prediction $\hat{x}$ to the ground truth trajectory $x$. Instead of learning $\mathcal{H}(\hat{x})$ directly, we attempt to train a residual module $\mathcal{F}(\hat{x}) = \mathcal{H}(\hat{x}) - \hat{x}$, which essentially learns the offset $\hat{o}$ to warp $\hat{x}$. In this way, the prediction from WLSTM is guaranteed to be no worse than the nominal prediction represented by an identity mapping.

In the residual module, a linear layer $\phi$ embeds $\hat{x}_{1:t+T_g}$ into $e_{1:t+T_g}$. A bidirectional LSTM is applied to encode the trajectory. Both the observation from the past and the intention information from the future are integrated into hidden states using LSTM along both forward and backward directions. The hidden states are decoded to offset output $\hat{o}_{1:t+T_g}$ by a linear layer $\gamma$. A skip connection is built to add the offset $\hat{o}_{1:t+T_g}$ to $\hat{x}_{1:t+T_g}$. Finally, WLSTM outputs $\hat{x}_{1:t+T_g}$, and $\hat{x}_{t+1:t+g}$ becomes the prediction. We define the loss function of WLSTM as the mean squared error of $\hat{x}_{1:t+T_g}$ from the ground truth $x_{1:t+T_g}$. Note the loss function includes the offset error through observation period in order to clip the observation part of warped trajectory towards the ground truth observation.






\begin{figure}[!t]
	\centering
	\includegraphics[width=50mm]{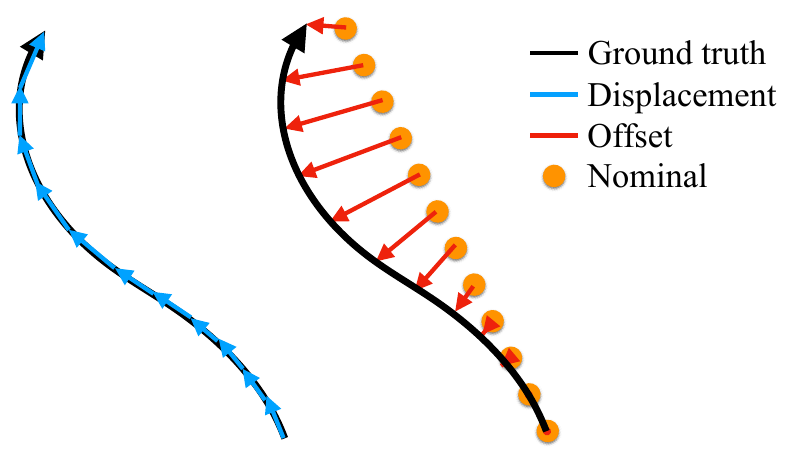}
	\caption{Displacement versus offset in trajectory prediction.}
	\label{fig:residual_module}
	\vspace{-10pt}
\end{figure}

\begin{equation}
\begin{aligned}
e_t &= \phi{\left(\hat{x}_t; W_e\right)} \\
\overrightarrow{h}_t &= \textrm{LSTM}\left(e_t, \overrightarrow{h}_{t-1}; \overrightarrow{W}\right) \\
\overleftarrow{h}_t &= \textrm{LSTM}\left(e_t, \overleftarrow{h}_{t+1}; \overleftarrow{W}\right) \\
\hat{o}_t &= \gamma(\overrightarrow{h}_t, \overleftarrow{h}_t; W_o) \\
\hat{x}_t &\gets \hat{x}_t + \hat{o}_t\\
\end{aligned}
\end{equation}






\section{Experiments}\label{sec:4}
\begin{figure}[hbt!]
	\centering
	\includegraphics[width=60mm]{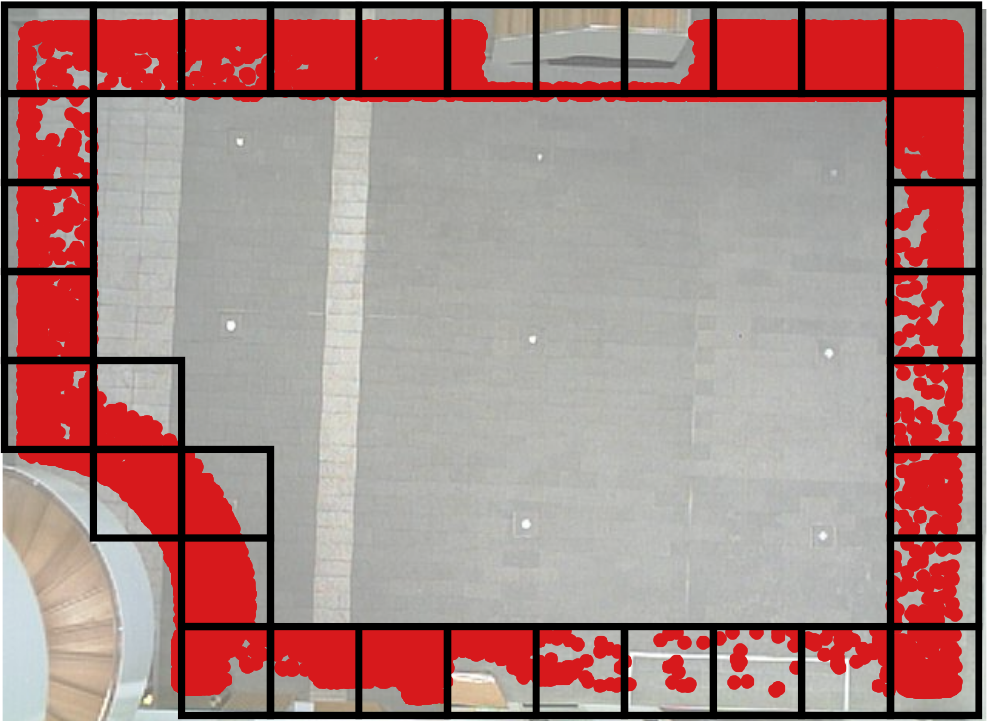}
	\caption{Visualization of goal positions (red dots) of the trajectories in Edinburgh trajectory dataset. 34 black squares with 1.5m $\times$ 1.5m size represent common goal regions and are defined as 34 potential intentions which belong to $\mathcal{G}$.}
    \label{fig:dataset}
    \vspace{-10pt}
\end{figure}

We conduct experiments on a publicly available dataset named Edinburgh Informatics Forum database \cite{majecka2009statistical}, which includes around 90,000 trajectories recorded at a frame rate of 10Hz. Fig. \ref{fig:dataset} shows that goal positions of full-length trajectories are clustered on the map boundaries, so 34 square-shaped goal regions along the edges are defined as possible human intentions. The pedestrian trajectories are split into training dataset (80\%) and test dataset (20\%). 

The experiment is composed of training and evaluation stages. During the training stage, WLSTM is trained using the training dataset, with the assumption that the ground truth goal position $g$ and the remaining time step $T_g$ are given. The performance of WLSTM during this stage is assessed across the remaining time period of full-length trajectories, so as to check the effective use of intention information. During the evaluation stage, MIF integrated with the trained WLSTM is applied to the test dataset. The observation history is available until the last observed time step, and the performance of the whole framework is evaluated with lookhead time window $T_f = 20$ \cite{fernando2018soft+}. The observation window is set as 20 for other baseline methods following the setting in \cite{fernando2018soft+}, since only a fixed-length observation sequence is used in these methods.

\subsection{Implementation Details and Metrics}
The embedding dimension is 128 in the residual module. The dimension of hidden states is 128 for each direction in LSTM. The Adam optimizer with an initial learning rate of 0.0001 is used to train WLSTM \cite{kingma2014adam}. The mutation probability $P_{mutation}$ is set to 0.01. The filtering process iterates per 2 time steps. The number of particles $M$ is 340. The trajectory prediction performance is quantified by the first four metrics below. The last metric is intended for evaluation on intention estimation. Lower values in the first four metrics are better, and a higher value in the last one is better.

\begin{enumerate}[1.]
    \item \textit{Average Offset Error (AOE)}: The average of L2 distance between the predicted and ground truth trajectories at each time step of the prediction period \cite{pellegrini2009you}.

    \item \textit{Final Offset Error (FOE)}: The L2 distance between the predicted and ground truth trajectories at the final time step of the prediction period.
    
    \item \textit{Max Offset Error (MOE)}: The max L2 distance between the predicted and ground truth trajectories across all time steps in the prediction period.
    
    \item \textit{Negative Log Likelihood (NLL)}: The average negative log likelihood of the ground truth trajectory given the kernel density estimate from all prediction samples \cite{ivanovic2019trajectron}. 
    
    \item \textit{Intention Estimation Accuracy (IEA)}: Percentage of correct intention estimate at the final iteration of filtering process. The estimation is defined as correct if one of top-probability intentions belongs to the set of the ground truth intention and its two adjacent intentions.
    
\end{enumerate}

\subsection{Comparison with Related Work}\label{sec:experiments-comparison}

We compare our results during the evaluation stage with several existing approaches. (1) Social Force (SF): a model-based method that includes preferred speed and neighbors' motion \cite{helbing1995social}; (2) Vanilla LSTM; (3) Social LSTM (SLSTM): A LSTM with a social pooling layer to model pedestrian interaction \cite{alahi2016social}; (4) Attention LSTM (ALSTM): An attention-based LSTM that aggregates information among pedestrian and neighbors across the whole observation period \cite{fernando2018soft+}; (5) Social GAN (SGAN): A socially-aware LSTM-based Generative Adversarial Network that generates multi-modal trajectory prediction samples \cite{gupta2018social}. 

\begin{figure}[t!]
	\centering
	\includegraphics[width=70mm]{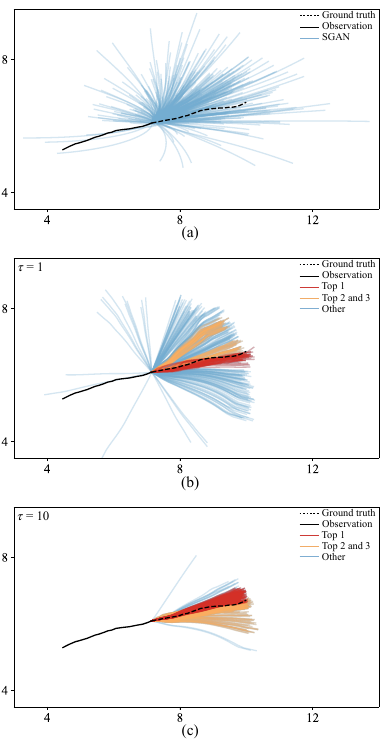}
	\caption{Prediction samples generated by (a) SGAN, (b) MIF-WLSTM$_{1,2,3}$, and (c) MIF-WLSTM$_{4,5,6}$.}
    \label{fig:visual_nll}
    \vspace{-15pt}
\end{figure}

The results are presented in Table \ref{table:baseline_comparison_eval}. For fair comparison, SGAN generates the same amount ($M=340$) of samples as MIF-ILSTM. The mean of AOE/FOE is computed to represent the average performance of prediction samples. One possible reason that our method outperforms SGAN is the variety loss used to train SGAN only considers the minimum L2 error over prediction samples. Fig. \ref{fig:visual_nll} illustrates that the variety loss encourages generation of diverse trajectories for SGAN, and SGAN works better than our method in terms of minimum AOE/FOE as presented in TABLE \ref{table:parametric_study}. However, SGAN does not provide a tool to differentiate the best prediction from others. Since all predictions are treated identically, in spite of the decent minimum error, the multi-modality may lead to conservative prediction of possible future trajectories and an unimpressive mean error. In contrast, the prediction samples of our method are the particles in MIF, and each of them is assigned an intention hypothesis $G^{(i)}$. As the belief $b_G$ after each iteration reveals intention hypotheses with the highest probabilities, a subset of prediction samples corresponding to these hopeful intention candidates can be regarded as more trustworthy than others, and may be readily selected in place of the whole set. Fig. \ref{fig:visual_nll} displays the subset of samples corresponding to intention hypotheses with maximum probability and top three probabilities in $b_G$, and demonstrates the flexibility of sampling strategies allowed by our framework. 

\begin{table}[t!]
\caption{Performance comparison during evaluation stage. Subscripts of our method denote ID 1 and 4 in TABLE \ref{table:parametric_study}, which represent the best performance with $\tau$ set as 1 and 10.}\label{table:baseline_comparison_eval}
\begin{center}
\begin{tabular}{lcc}
    \toprule
    Method & AOE & FOE \\
    \midrule
    SF&	3.124&	3.909 \\
    LSTM&	2.132&	3.005 \\
    SLSTM&	1.524&	2.510 \\
    ALSTM&	0.986&	1.311 \\
    SGAN&	1.042&	2.088 \\
    MIF-WLSTM$_1$& 0.665&	1.236 \\
    MIF-WLSTM$_4$& \textbf{0.636}&	\textbf{1.179} \\
    \bottomrule
\end{tabular}
\end{center}
\vspace{-1pt}
\end{table}

\begin{table*}[hbt!]
\caption{Parametric study on MIF-WLSTM. $\tau$ is the temperature hyperparameter in MIF. IMM denotes intention mutation mechanism. NTI denotes the number of top-probability intentions. Intention Estimation Accuracy (IEA), Negative Log Likelihood (NLL), mininum and mean Average Offset Error/Final Offset Error (AOE/FOE) are reported.}\label{table:parametric_study}
\begin{center}
\begin{tabular}{l|c|ccc|cccccc}
    \toprule
    Method & ID & $\tau$ & IMM & NTI & IEA & NLL & min AOE & min FOE & mean AOE & mean FOE \\
    \midrule
    \midrule
    SGAN& N/A& & N/A& & N/A & 5.153&	\textbf{0.151} & \textbf{0.178}& 1.042 & 2.088  \\
    \midrule
    
    \multirow{12}{*}{MIF-WLSTM} & 1& 1&	\checkmark&	1&	75.0\%& 7.773&	0.448&	0.774&	0.665&	1.236 \\
    & 2& 1&	\checkmark&	3&	92.3\%& 4.411&	0.298&	0.455&	0.686&	1.279 \\
    & 3& 1&	\checkmark&	34&	N/A& \textbf{2.412}&	0.221&	0.283&	0.778&	1.463 \\
    
    & 4& 10 & \checkmark & 1& 77.2\%& 	7.717&	0.411&	0.698&	\textbf{0.636}&	\textbf{1.179} \\
    & 5& 10 & \checkmark & 3 & 89.0\%&  5.664&	0.315&	0.493&	0.640&	1.187 \\
    & 6& 10 & \checkmark & 34&N/A&  4.689&	0.282&	0.421&	0.650&	1.206\\
    & 7& 1&	-&	1& 73.0\%& 	7.970&	0.494&	0.865&	0.711&	1.325 \\
    & 8& 1&	-&	3& 88.4\%& 	5.035&	0.367&	0.591&	0.732&	1.368 \\
    & 9& 1&	-&	34&	N/A& 3.821&	0.318&	0.478&	0.813&	1.529 \\    
    & 10& 10&	-&	1& 62.6\%& 	8.699&	0.564&	1.001&	0.790&	1.479 \\
    & 11& 10&	-&	3&	79.6\%& 7.455&	0.514&	0.891&	0.793&	1.484 \\
    & 12& 10&	-&	34&	N/A& 7.168&	0.503& 0.866&	0.795&	1.489 \\
    \bottomrule
    
\end{tabular}
\end{center}
\vspace{-10pt}
\end{table*}

\subsection{Parametric Study}

Fig. \ref{fig:visual_nll} not only presents the difference between SGAN and our method, but also visualizes the influence of different components of our method on trajectory prediction. Specifically, We analyze how number of top-probability intentions (NTI), weight update hyperparameter $\tau$, intention mutation mechanism (IMM), and motion models affect the performance.

Lower NTI represents more aggressive sampling strategies. As Fig. \ref{fig:visual_nll}(b)(c) have shown, the red samples (NTI = 1) are more concentrated than red and orange samples together (NTI = 3), let alone the whole set. On the one hand, a more concentration subset implies a better average prediction performance, since samples in the subset are often more aligned with the desired moving direction. Choosing only the samples with the maximum-probability intention outperforms choosing the whole set by 7.5/7.9(\%) in terms of mean AOE/FOE. On the other hand, choosing higher NTI would guarantee a better min AOE/FOE, because the sample subset of a lower NTI is a subset of the counterpart of a higher one.

The weight update in MIF is controlled by $\tau$. A higher $\tau$ amounts to faster convergence to the most probable intention hypotheses. Thus, we see in Fig. \ref{fig:visual_nll} that the whole set of prediction samples are more concentrated when $\tau = 10$ against when $\tau = 1$. While both higher NTI and higher $\tau$ result in concentration, the former achieves this goal by picking a subset from a potentially high multi-modal set of samples, and the latter attempts to enforce uni-modality onto the whole set. The trend of uni-modality would easily give rise to deviation between ground truth trajectory and the set of prediction samples, and thus a poor result on NLL that evaluate a set of samples in a distributional manner. For example, TABLE \ref{table:parametric_study} shows that NLL of the variant $6$ with $\tau = 10$ is 94.4\% larger than the variant $3$ with $\tau = 1$ when the whole set of samples is considered.

\begin{figure}[t!]
	\centering
	\includegraphics[width=70mm]{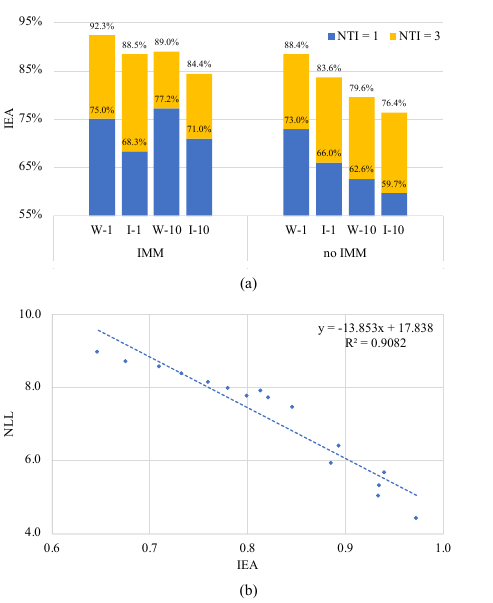}
	\caption{(a) IEA given different variants of our method. The configuration pair on X axis is motion model-$\tau$, where W is WLSTM and I is ILM. (b) Linear regression between NLL and IEA among all variants of MIF-WLSTM and MIF-ILM.}
    \label{fig:stats_imm_nl_iea}
    \vspace{-15pt}
\end{figure}

Fig. \ref{fig:stats_imm_nl_iea}(a) illustrates that IMM is essential to boost the intention estimation performance of MIF. IEA is improved by 10.1\% in average among all configurations by adding IMM to the filtering process of MIF. In particular, the variant 4 outperforms the variant 6 in TABLE \ref{table:parametric_study} by 23.3\%, since $\tau = 10$ forces particles to quickly converge to a single intention hypothesis, which is dangerous without IMM as the particles with other intention hypotheses are extinct and unable to be recovered.

As for motion models, both ILM and WLSTM are integrated with MIF for the evaluation stage. Fig. \ref{fig:stats_imm_nl_iea}(a) reveals that MIF-WLSTM works better than MIF-ILM in terms of IEA, and suggests a better motion model indeed enhance the performance of intention estimation. Moreover, we conducted linear regression between NLL and IEA and found they are highly correlated, which indicates that trajectory prediction and intention estimation are highly dependent on each other.

\subsection{Intention Information Extraction}

We found that during the training stage, it is more efficient to train WLSTM on full trajectories with a fixed percentage observed rather than using a random percentage as observation. Thus, we choose four representative percentages of trajectories (0\%\footnote{0\% observation is equivalent as the position and the displacement at the first time step.}, 25\%, 50\%, 75\%) to split a full-length trajectory into observation and prediction in order to investigate prediction algorithms at different stages along the trajectory. One of these models is selected based on the ratio between the length of observation history and the remaining time step $T_g^{(i)}$ estimated in Algorithm \ref{algo:ITP} during evaluation stage.

\begin{figure}[t!]
	\centering
	\includegraphics[width=70mm]{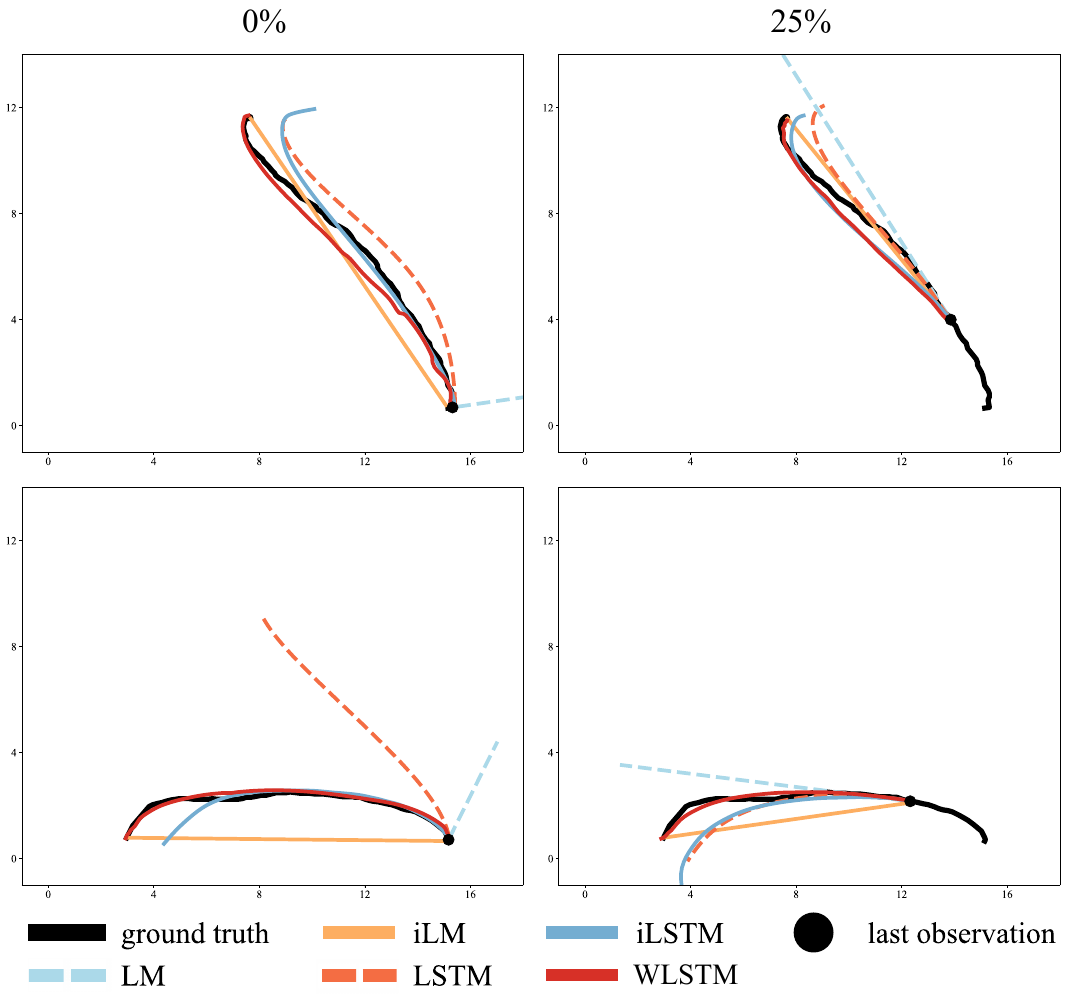}
	\caption{Full trajectory prediction during training stage.}
    \label{fig:traj_vis}
    \vspace{-15pt}
\end{figure}

In addition to baselines including linear model (LM), our nominal model ILM, and vanilla LSTM, an intention-aware LSTM is introduced to represent the existing methodology of integrating intention information into a unidirectional LSTM~\cite{rasouli2019pie}. Basically, the ground truth goal position $g$ will be input along with the displacement at current time step to LSTM in order to predict the displacement at next time step. TABLE \ref{table:baseline_comparison_training} discloses that though ILSTM is fed with the $g$ at each time step, its FOE, which means the offset error between predicted goal position and $g$, is still comparable to MOE. This implies ILSTM still suffers from the drifting issue caused by sequential displacement prediction just like vanilla LSTM. As bottom-left subfigure in Fig. \ref{fig:traj_vis} suggests, ILSTM could learn from $g$ a roughly correct moving direction compared to LSTM when observation is insufficient. In contrast, our WLSTM makes full use of coordinate information of intention, by effectively clipping prediction at final time step to $g$ based on the nominal prediction from ILM, and learns the pedestrian motion pattern to ``warp" the nominal trajectory.

\begin{table}[t!]
\caption{Performance comparison during training stage.}\label{table:baseline_comparison_training}
\begin{center}
\begin{tabular}{lccc}
    \toprule
    Method & AOE & FOE & MOE \\
    \midrule
    LM	&2.401	&4.987	&5.044 \\
    ILM	&0.580	&N/A	&1.040 \\
    LSTM	&1.778	&3.459	&3.570 \\
    ILSTM	&1.394	&2.440	&2.604 \\
    WLSTM	&\textbf{0.427}	&\textbf{0.201}	&\textbf{0.792} \\
    \bottomrule
\end{tabular}
\end{center}
\vspace{-10pt}
\end{table}

\subsection{Intention-changing Scenarios}

\begin{figure}[t!]
	\centering
	\includegraphics[width=75mm]{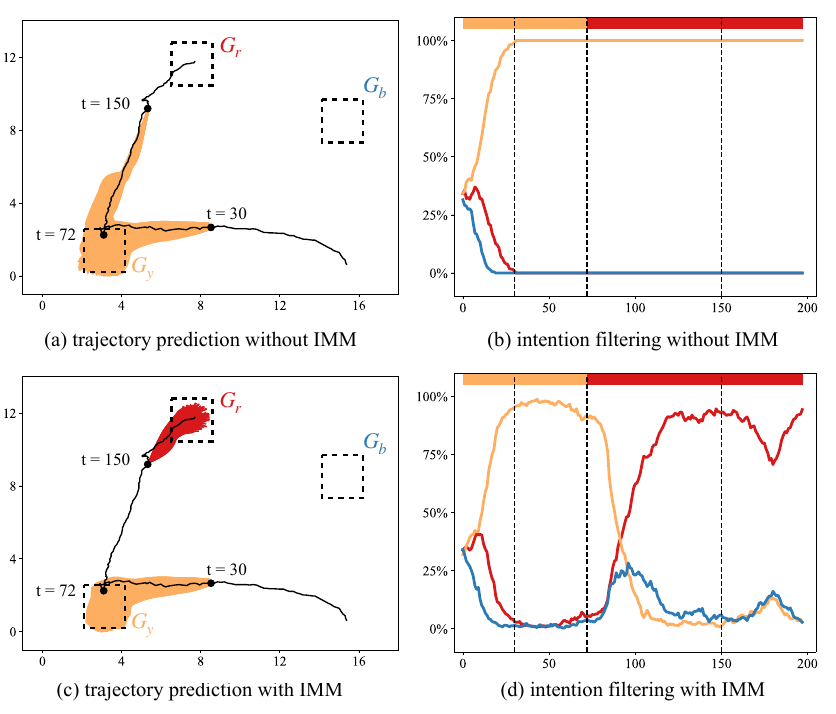}
	\caption{Visualization of filtering process on an intention-changing trajectory. At the beginning, the pedestrian walked from bottom-right corner to the intention $G_y$. At $t = 72$, the intention was changed to $G_r$. NTI is set as 1, and only three potential intentions are considered for the convenience of illustration. (a) and (c) present full trajectory prediction, where black represents ground truth and other different colors correspond to prediction with different intention hypotheses. (b) and (d) present the intention probability distribution throughout the filtering process. The bars at top of (b) and (d) correspond to intention annotations of the trajectory, while the dashed lines are the time steps marked on (a) and (c).}
    \label{fig:vis_mutation_no_mutation}
    \vspace{-15pt}
\end{figure}

We further investigate the capability of our work on intention-changing pedestrian trajectories. From Fig. \ref{fig:vis_mutation_no_mutation}(b), we see that without IMM, particles quickly converges to $G_y$, and probabilities of $G_r$ and $G_b$ dropped to zero and never grow back. Thus, the prediction samples at $t=150$ are still directed towards $G_y$. In contrary, IMM in MIF guarantees nonzero probability for $G_r$ and $G_b$ as shown in Fig. \ref{fig:vis_mutation_no_mutation}(d), while the existence of particles with $G_r$ and $G_b$ as intention hypotheses does not affect trajectory prediction at $t = 30$ in Fig. \ref{fig:vis_mutation_no_mutation}(c). Thus, MIF can react promptly once the pedestrian exhibits the motion towards $G_r$, and switch the maximum-probability intention to $G_r$.

\section{Conclusions}\label{sec:5}

In this work, we present a novel trajectory prediction framework to model long-term pedestrian behavior. We propose a Mutable Intention Filter enabling robust intention estimation even for intention-changing pedestrian motion. A Warp LSTM is introduced to effectively extract intention information by bidirectional propagation through both observation and nominal prediction, and to predict offset over the full trajectory that resolves the drifting prediction issue caused by sequential displacement prediction. We demonstrate that our framework surpasses baselines on a publicly available dataset. In the future we would like to integrate our current study on global-scale goal-directed motion with local-scale human-human interaction within a unified framework, and implement the framework in human-robot interaction tasks.

\bibliographystyle{IEEEtran}
\bibliography{strings, bib}

\end{document}